\pdfoutput=1
\documentclass[11pt]{article}

\usepackage[]{acl}

\usepackage{times}
\usepackage{latexsym}

\usepackage[T1]{fontenc}
\usepackage[utf8]{inputenc}

\usepackage{microtype}
\usepackage{booktabs}
\usepackage{tabularx}
\usepackage{colortbl}
\usepackage{color}

\usepackage{graphicx}
\graphicspath{ {./images/} }

\definecolor{light-gray}{gray}{0.9}

\title{Text Simplification of College Admissions Instructions:\\A Professionally Simplified and Verified Corpus}

\author{Zachary W.\ Taylor \\
  College of Education and Human Sciences\\
  University of Southern Mississippi \\
  \texttt{z.w.taylor@usm.edu} \\\And
  Maximus H.\ Chu \\
  Department of Computer Science \\
  The University of Texas at Austin \\
  \texttt{maximuschu@utexas.edu} \\\AND
  Junyi Jessy Li \\
  Department of Linguistics\\
  The University of Texas at Austin \\
  \texttt{jessy@utexas.edu}
  }

\begin{document}
\maketitle
\begin{abstract}
Access to higher education is critical for minority populations and emergent bilingual students. However, the language used by higher education institutions to communicate with prospective students is often too complex; concretely, many institutions in the US publish admissions application instructions far \emph{above} the average reading level of a typical high school graduate, often near the 13th or 14th grade level. This leads to an unnecessary barrier between students and access to higher education.

This work aims to tackle this challenge via text simplification. We present \textbf{PSAT} (\textbf{P}rofessionally \textbf{S}implified \textbf{A}dmissions \textbf{T}exts), a dataset with 112 admissions instructions randomly selected from higher education institutions across the US. These texts are then professionally simplified, and verified and accepted by subject-matter experts who are full-time employees in admissions offices at various institutions. Additionally, PSAT comes with manual alignments of 1,883 original-simplified sentence pairs. The result is a first-of-its-kind corpus for the evaluation and fine-tuning of text simplification systems in a high-stakes genre distinct from existing simplification resources. PSAT is available at \url{https://doi.org/10.5281/zenodo.7055024}.
\end{abstract}

\section{Introduction}

Access to the higher education system for minoritized populations, especially low-income students, first-generation in college students, students of Color, and students whose first spoken language is not English, is an important social challenge \cite{auerbach2004engaging, cook2012increasing, flores2010state, perez2008journal, rosa2006opportunity,taylor2018intelligibility, taylor2020comprenderan}. 
However,
researchers have consistently explained that much of higher education's communication with prospective students is too complex, too lengthy, and requires a wealth of prior knowledge of the higher education system to successfully navigate \cite{auerbach2004engaging, perez2008journal}; 
concretely many institutions in the US publish admissions application instructions at or above the 14th grade English reading comprehension level, far too high for the average prospective student or adult to read and comprehend \cite{taylor2018intelligibility, taylor2019lost,taylor2020college}.
As a result, 
the verbose, difficult communication places an unnecessary barrier between students and access to higher education \cite{ardoin2013learning, goff2004preferred, hartman1997college, kanno2018non, taylor2017twenty}.

With modern conditional text generation models, one plausible way to lower such barriers is automatic text simplification~\cite{siddharthan2014survey,alva2020data}, i.e., simplify text such that it is more readable and accessible, while adhering to the texts' original semantic content. Text simplification is known to benefit a range of readers, including children~\cite{javourey2022simplification} and L2 learners of English~\cite{yano1994effects}. Yet, no studies have explored how to simplify this information, without losing important details, to render the admissions process more accessible for prospective students and their families. Crucially, existing work in text simplification largely rely on two established datasets in the News and Wiki domains~\cite{woodsend2011wikisimple,coster2011simple,xu2015problems,zhang2017sentence}; unlike these two domains, college admission texts certainly falls under the category of being more specialized and difficult to generalize, where both the jargon and concepts encapsulated within these texts are not ones that one encounters every day. Consequently we lack understanding both in terms of the nature of simplification in college admissions texts, as well as model performance in this new domain. The lack of domain diversity has been identified as a critical issue of text simplification as a whole~\cite{alva2020data}.

\begin{table}[t]
    \centering
    \small
    \begin{tabular}{p{7cm}}
    \toprule
    \textbf{(Original)} All conditionally accepted applicants must consent to, submit to and successfully complete a criminal background check through Certiphi Screening, Inc. Failure to do so will constitute failure to meet the pre-matriculation requirements established by SUNY Optometry and will result in the withdrawal of a conditionally accepted offer. \\
    \rowcolor{light-gray}
    \textbf{(Simplified)} If you are conditionally accepted, you must consent to, submit to, and complete a criminal background check through Certiphi Screening, Inc. If you do not, we will withdraw your conditionally accepted offer. \\
    \midrule
    \textbf{(Original)} You must complete the following steps before USF will consider your application complete and begin admission evaluation. Pay the non-refundable \$30 application fee or submit an application fee waiver. \\
    \rowcolor{light-gray}
    \textbf{(Simplified)} You must submit an online application with a nonrefundable \$30 application fee. You can also submit an application fee waiver.\\
    \bottomrule
    \end{tabular}
    \caption{Original vs our expert simplified version of college admissions texts.}
    \label{tab:introexamples}
\end{table}

This work introduces \textbf{PSAT} (\textbf{P}rofessionally \textbf{S}implified \textbf{A}dmissions \textbf{T}exts), the first manually simplified and verified corpus of admissions instructions from 112 randomly sampled US post-secondary institutions.
Our professional simplification process are guided by extensive existing literature on manual simplification, including lowering syntactic complexity, improving lexical cohesion, and elaborating jargon~\cite{crossley2008text,mcnamara2014automated,siddharthan2014survey}. 
Given that the admissions process normally requires application fees, completion of an application, writing and submitting of essays, sending of transcripts, and possibly more processes, it is crucial to understand how to compose admissions application instructions to clearly but accurately explain these processes, especially to first-generation in college students who may not have support from their secondary school, family, or household \cite{taylor2020college}.
\emph{To this end, every document in PSAT was manually verified by 2 subject-matter experts (among a total of 10 experts) who are employed as full-time admissions professionals in US institutons of higher education}. 

In addition to full document simplification, we also create manual alignments (by the author who originally simplified all documents), aligning each sentence in the original text to its simplified version. This entailed 1,883 sentence pairs total.

Our analyses showed that the simplified documents reduced the reading level of these texts from grade 13.3 to grade 9.8, making it much more accessible for minority and emergent bilingual students. 
For experiments on automatic simplification, in this paper we focus on \emph{sentence} simplification utilizing the  aligned sentence pairs. We start from zero-shot transfer from existing simplification models pre-trained on both Wikilarge~\cite{zhang2017sentence} and Newsela~\cite{xu2015problems}, as well as fine-tuned models on PSAT. We showed that this is a challenging domain for simplification. Additionally, our professional manual inspection of model outputs points out domain-specific errors that impact the accuracy of the simplifications.

\section{Background and related work}

\subsection{Readability in admissions text}
Decades of research in higher education has demonstrated that higher education information is often unreadable by prospective student audiences who read at average, pre-enrollment reading comprehension levels, roughly the 12th grade \cite{taylor2018intelligibility, taylor2019lost, taylor2020college}. At a larger scale, literacy research has consistently found that the average United States adult reads and comprehends below the 9th grade level, suggesting that many adults many not be able to understand how to read college admissions information and successfully apply to an institution of higher education in the United States \cite{hauser2005measuring, mamedova2019adult, sum2004pathways}.

Pertinent to the study at hand, prior research has found that undergraduate admissions instructions (what a student needs to accomplish and submit to the institution to be considered for admission) are often written above the 14th grade English reading comprehension level \cite{taylor2020college}. Moreover, qualitative research focused on student experiences during the application process has found that prospective students, many of them from low-income and first-generation in college backgrounds, often struggle understanding higher education jargon such as \textit{undergraduate}, \textit{FAFSA}, and \textit{verification}, all of which are critical to comprehend during the admissions and enrollment management process \cite{ardoin2013learning, taylor2019fafsa}. Here, many students may be academically prepared for the rigors of a college curriculum, yet they may not understand the language and processes of higher education writ large. Such a misunderstanding may systematically exclude these students due to a lack of support during the application process, not a lack of preparedness for college itself.

However, little has changed in the decades during which higher education researchers have been investigating the complexity of the college admissions process. In the 1980s and early 1990s, researchers called for the admissions process to be simplified and for common admissions applications to be widely embraced by institutions across the country \cite {astin1982minorities, gandara1986chicanos, post1990college}. Yet by the early 2000s, researchers were still reporting that many minoritized students and their families struggled to comprehend how to complete the college admissions process and enroll in higher education \cite{mcbrien2005educational, rosa2006opportunity, tornatzky2002college, ward2006improving}. 

Recently, given the considerable personal and financial hardships facilitated by the COVID-19 pandemic, students and families continue to report that the college admissions process is too difficult \cite{mcculloh2022exploration} and often requires dozens of hours to complete \cite{reilly_2021}, and that institutions of higher education have not made efforts to ease the admissions process by simplifying the instructions for how to apply and enroll into higher education \cite{hurtado2020latinx, morrison_2021}. Subsequently, this study attempts to analyze simplifications of college admissions instructions to create an automated model that may automatically simplify college admissions instructions, easing the information burden for prospective students and families, while also easing the workload of admissions practitioners working for institutions of higher education.

\subsection{Text simplification}
Modern text simplification systems are typically encoder-decoder models that performs monolingual translation from the original to simplified text~\cite{wang2016text,zhang2017sentence,kriz2019complexity,dong2019editnts,martin2020controllable,devaraj2021paragraph,maddela2021controllable}. The majority of this research are at the sentence level, trained using two corpora: the Wikipedia-Simple Wikipedia aligned corpus~\cite{woodsend2011wikisimple,coster2011simple,zhang2017sentence} and the Newsela simplification corpus~\cite{xu2015problems}. As a result, it is unclear whether models trained on these datasets can be applied in other domains, and progress is hindered by the lack of diverse datasets~\cite{alva2020data}. This work is a step towards mitigating towards this issue.

The training of supervised models rely on automatically aligned sentence pairs~\cite{jiang2020neural}. However, \citet{devaraj2022evaluating} demonstrated that noisy alignments can lead to inaccuracies during the simplification process. Instead, in this work, all documents are manually aligned by the author of the simplification.

\section{Data collection and simplification}

The main work of PSAT included performing manual text simplification, followed by acceptability judgements made by subject-matter experts with domain-specific professional work experience in either admissions or financial aid. Below, we detail the corpus development process, including the gathering of college admissions instructions (Section~\ref{sec:rawtext}), professional simplification and its principles (Section~\ref{sec:simplification_process}) and verification (Section~\ref{sec:sme}).

\subsection{Raw text extraction}\label{sec:rawtext}
To gather the original texts (i.e., publicly available admissions instructions texts) for this study, we employed the National Center for Education Statistics' Integrated Postsecondary Education Data System (IPEDS)\footnote{\url{https://nces.ed.gov/ipeds/}} to gather institutional URLs to each institution’s admissions application instructions website. 
Moreover, we decided to gather data during the college search and exploration process, typically occurring between August and November of each year in the US \cite{hossler1987studying}.
Understanding both student exploration and institutional information practices, we gathered all website data for this study in October 2019. Initially, we gathered  undergraduate admissions instructions from 335 institutions, which we then manually inspected and kept 112 that contain long, meaningful discourse other than metadata (e.g., addresses).

All instructions are then cleaned, by manually extracting the raw text.

\subsection{Simplification}\label{sec:simplification_process}

\paragraph{Personnel} The texts, written in English (from US institutions), were manually simplified by an author of this paper who is a native English speaker. The author has a doctoral degree in education and has worked professionally in US postsecondary education for over a decade, including work in undergraduate admissions.
Thus the author engaged with their professional insight to 
simplify without losing critical information necessary for its comprehension and understanding. 

\paragraph{Principles} 
Longitudinal research~\cite{crossley2008text,mcnamara2014automated,siddharthan2014survey} has suggested that text can often be simplified using a smaller lexicon than the original text, rewriting sentences so that adjoining sentences share syntactic features (e.g., punctuation marks, independent clauses followed by main clauses), and ensuring that previously used lexical items appearing earlier in a text are re-introduced later in the text to improve comprehension of the text. Thus our manual simplification process include \emph{reducing syntactic complexity}, 
\emph{increasing lexical cohesion} (word overlap and frequency),
\emph{the elaboration and explanation of jargon and acronyms}, and \emph{domain-specific principles}.

\textbf{\emph{(1) Reducing syntactic complexity.}}
We adopted \citet{coleman1962improving}'s framework for sentence-level simplification: raising clause fragments to full sentences, dividing sentences joined by conjunctions (e.g., because, but, for, or), avoid dividing sentences joined by the conjunction ``and'', and shortening clauses by using periods where other forms of grammatical punctuation may be found (e.g., semicolons, colons, commas). While shortening, we ensured that we do not lose critical information which would cause a factual error~\cite{devaraj2022evaluating}.

Additionally, extant research has suggested that writing or speaking in active voice rather than passive voice can increase simplicity, and this rewrite can often lead to shorter sentences \cite{devito1969some, ferreira1994choice}. As a result, we identified instances of passive voice in each admissions text and re-wrote them into active voice, e.g.,
\begin{quote}
\small
    \textbf{(Original)} The application must be completed by the student.
    
    \textbf{(Simplified)} The student must complete the application.
\end{quote}

\textbf{\emph{(2) Increasing lexical cohesion.}} \citet{hulme1997word} learned that increasing the word frequency in an informative text helped with the short-term memory recall of research participants regarding the content of the text, supporting the finding that increasing the word frequency in a text may lead to a better understanding of the text on behalf of the reader. These words included "you," "must," "official," "transcript," and many others. Therefore, we simplified in a way that promotes lexical overlap across documents.

Pertinent to this study, \citet {monaghan2017exploring} also found that individual differences across bilingual readers in terms of word frequency effects were due to exposure to word diversity, not an individual's vocabulary size (personal lexicon). This finding supported the use of increasing word frequency to increase a text’s simplicity and possible readers' comprehension of the text. Thus, we also attempted to identify content words that could be repeated earlier or later in each text separately, including words like "you," "must," "submit," "transcripts," "contact," and many others.

\textbf{\emph{(3) Elaboration and explanation of jargon and acronyms.}}
Research on acronyms and initialisms has found that using these lexical items in potentially unfamiliar text can be confusing to readers, thus making the text more difficult to read \cite{cannon1989abbreviations, ibrahim1989acronyms, rua2002structure,laszlo2007better}. Using acronyms often hinders clarity, as the reader may need to parse extra text or consult another text in order to decipher the acronym or initialism and fully comprehend the text \cite{rua2002structure, taghva1999recognizing}. 
We attempted to locate acronyms (e.g., ACT) and initialisms (e.g., GPA) within each text and ensure that these acronyms and initialisms were clear and commonly used, so that students would not be confused when reading the simplification. Common acronyms and initialisms in this dataset included "ACT," "SAT," "GED," "US," and several others. Subject-matter experts determined that although acronyms and initialisms may be confusing, including the acronyms and initialisms without lengthy definitions was best to assist with simplification, as subject-matter experts determined that the vast majority of students would understand these acronyms and initialisms without context clues or definitions.

\textbf{\emph{(4) Domain-specific principles.}}
As admission instructions are tied with direct consequences for postsecondary education, we work with 10 subject-matter experts (SME, their background detailed in Section~\ref{sec:sme}) such that our simplification is accurate as possible within this particular domain.

Subject-matter experts unanimously determined that admissions instructions should only contain the instructions themselves, and that any extraneous text should be removed to ensure minimalism and simplicity. Consequently,
%In addition, 
text was removed (from the original instructions) by SMEs if the text was not pertinent to the application process itself, including welcome statements, espoused institutional beliefs, and language that was determined to be marketing and/or branding and not admissions instructions.
Examples of removed texts include, \textit{``You don't have any time to waste, so we made applying as simple as possible''},
\textit{``The Office of Admissions assists prospective students in exploring the academic opportunities''}.

Moreover, there were several keywords within sentences that subject-matter experts insisted remain or be added from original to simplified texts. For example, the words ``official'' and ``transcript'' needed to be retained in the simplified versions; and if these words did not appear in the original, they were added into the simplified version for clarity for the prospective student. In addition, SMEs agreed that simple admissions text includes second-person pronoun usage and modal verbs to provoke action on behalf of the student, for example:
\begin{quote}
\small
\textbf{(Original)} Application fee waivers are available for students with demonstrated financial need.

\textbf{(Simplified)} If you cannot pay the fee, you can apply for an application fee waiver.
\end{quote}
Ultimately, subject-matter experts felt that this language rendered the text simpler because students would better understand that they were responsible for completing parts of the admissions process.

\subsection{Acceptability Judgements by Subject-Matter Experts (SMEs)}\label{sec:sme}

To determine whether the simplification of admissions application instructions were acceptable---that is to say they did not lose critical information or accuracy between the pre- and post-simplification process—we engaged with ten subject-matter experts (SMEs). Each simplified text was verified by 2 SMEs independently; in total, we engaged with 10 SMEs, who volunteered their time.

\paragraph{Personnel} All ten of the SMEs had professional backgrounds in U.S. postsecondary admissions, having worked at least five years full-time in college admissions offices in the United States. These SMEs were identified through professional networks and snowball methods, as several of our SMEs knew colleagues from different institutions or educational entities who would serve as high-quality, knowledgable SMEs. 

Moreover, we engaged with a diverse group of SMEs from different institution types (i.e., community colleges, public four-year institutions, private liberal arts colleges) and with various lengths of experience to capture the potential variability of admissions and financial aid parlance, jargon, and communication style. As the first study of its kind, identifying SMEs from diverse backgrounds provided more generalizability and reliability of findings, thus informing future research and practice regarding the communication of admissions application instructions to  students and their support networks. Four subject-matter experts worked at public, four-year universities, four worked at private, four-year universities, and two worked at public, two-year community colleges.

\paragraph{Process} To perform the acceptability judgement, the SME was presented with both pre- and post-simplification texts in real time over a Zoom video conference meeting. Then, we asked the SME to read the pre-simplified (original) text, followed by the post-simplification (simplified) text and determine whether the simplified text was acceptable. For example, changing the verb ``submit'' to ``complete'' is not acceptable because ``submit'' implies the documentation or information is being submitted by a submitter to a submittee, while ``complete'' only implies the documentation or information is completed and not directed to any educational stakeholder.

If a simplification was deemed unacceptable by one or more SMEs, we asked the SME what simplification would be acceptable through an iterative process in real time across all texts in this study. 
Once the SME provided their feedback and we integrated their feedback into the simplified text, the same SME was again asked to read both the both pre- and post-simplification texts in real time and render their acceptability judgement. If at any time there was an instance where a lexical item (e.g., single word, acronym, initialism, compound adjective), sentence, or paragraph could not be simplified, the pre-simplified section of that text was used.

\subsection{Manual Alignments}\label{sec:alignments}
In the PSAT corpus, the author of the simplified documents manually added sentence alignments across the original and simplified texts. Specifically, for each sentence in the original version, we provide the indice(s) of the corresponding simplified sentence. This comes to a total of 1,883 alignments, excluding sentences that are kept unchanged, metadata (e.g., addresses), and short ($<2$-word) headers.

\section{Analysis}\label{sec:stats}
Table~\ref{tab:stats} shows comparative statistics of the 112 original---simplified bitexts; we see that on average the simplified versions have shorter and fewer sentences. The original texts in PSAT are above the 12-th grade reading level (measured with Flesch-Kincaid~\cite{kincaid1975derivation}) of an average high school graduate, confirming the findings in~\citet{taylor2020comprenderan,taylor2020college}. The simplified version lowered this reading barrier.

We also study lexical items that are most associated with the original texts vs.\ the simplified version. Concretely we calculate the log-odds ratio of each word $w_i$, comparing the conditional probability of $w_i$ in the original instructions $\mathbf{D}_o$ or the simplified ones $\mathbf{D}_s$: logodds$(w_i)=\log(P(w_i|\mathbf{D}_o)/P(w_i|\mathbf{D}_s))$~\cite{nye2015identification}. Excluding punctuation and numbers, the words with the highest and lowest log-odds ratios with respect to simplified texts are shown in Table~\ref{tab:logodds}. We see that after simplification, the vocabulary becomes more standardized towards the application and admissions process itself.

\begin{table}[t]
    \centering
    \small
    \begin{tabular}{p{7cm}}
    \toprule
    \textbf{(Simplified texts)} just, development, following, prior, credit, sent, steps, recommended, applications, freshmen, payment, better, no, currently, once, chapman, plan, well, way, submitting, educational, aid, ensure, enroll, stonehill, options, completed, documents, code, create, credits, rolling, option, begin, appropriate, completing, now, earned, out, reviewed, checklist, items, submission, during, process, general, colleges, receive, please, standardized \\
    \midrule
    \textbf{(Original texts)} codes, materials, how, andor, csu, nonnative, paintings, prepare, invite, dualcredit, homeschool, cannot, using, words, admit, certiphi, can, must, native, responses, resume, you, selfreport, letter, georgia, closer, bonaventure, prompted, event, inclusion, syllabi, statements, dualdegree, significant, quebec, tb, former, thinking, filmmaking, carrolls, concise, testingservicesucmoedu, baton, crucial, parker, coach, cresson, did, esl, autobiography \\
    \bottomrule
    \end{tabular}
    \caption{Words most associated with simplified vs.\ original texts, calculated with log-odds ratios.}
    \label{tab:logodds}
\end{table}

\begin{table}[t]
\centering
\small
\begin{tabular}{llll}
\toprule
           & \# sents & \# tokens & FK \\ \midrule
Original   & 29.6    & 10.2 & 13.3      \\
Simplified & 17.9    & 7.1 & 9.8   \\
\bottomrule
\end{tabular}
\caption{Comparison of original vs.\ simplified texts: average number of sentences, tokens, and the Flesch-Kincaid grade level.}
\label{tab:stats}
\end{table}

Finally, we measure the abstractiveness of the simplified texts; namely, how much paraphrasing was used? For this we use the \emph{aligned} portion of PSAT; the percentage of unique uni-, bi-, and tri-grams in the simplified texts are 34.4\%,52.4\%,58.6\%, respectively. 
Thus the amount of new jargon introduced through simplification is enough to be significant; this indicates that simpler vocabulary is being introduced, and significant paraphrasing happened during simplification.

\section{Sentence Simplification Experiments}

We establish baselines for the sentence simplification task in PSAT, using the aligned sentences. With these experiments we hope to evaluate existing models---which are trained in other domains---in a zero-shot manner, gauge the utility of fine-tuning on PSAT, and discuss the challenges of simplification in this domain.

\subsection{Models}
\paragraph{ACCESS} Our baseline is the \texttt{ACCESS} model~\cite{martin2020controllable}, trained on Wikilarge~\cite{zhang2017sentence}. \texttt{ACCESS} uses a transformer trained from scratch (i.e., randomly initialized).

\paragraph{T5} We also fine-tune the T5 model~\cite{raffel2020exploring} for the simplification task. T5 is a pre-trained large-scale encoder-decoder model optimized on conditional generation tasks (e.g., summarization and machine translation), in addition to unsupervised objectives. We fine-tune T5-base on both Wikilarge and Newsela~\cite{xu2015problems}, denoted as \texttt{T5-wiki} and \texttt{T5-newsela}.
Following~\citet{devaraj2022evaluating}, the models are fine-tuned with prefix \emph{``summarize:''} for 5 epochs with a batch size of 6 and learning rate 3e-4. The default greedy decoding was used.

Additionally, we further fine-tune these models on PSAT's training data (Section~\ref{sec:setup}) for 3 more epochs, denoted as \texttt{T5-wiki-ft} and \texttt{T5-newsela-ft}.

\subsection{Experimental setup}\label{sec:setup}

\begin{table}[t]
\small
\centering
\begin{tabular}{llll}
\toprule
 & Train & Test & Validate   \\
\midrule
\# Files & 56 & 33 & 23   \\ 
\# Sentence Pairs & 955 & 559 & 369  \\ 
\bottomrule
\end{tabular}
    \caption{The number of files and sentence pairs for the train, test, validation splits in PSAT.}
    \label{tab:data}
\end{table}

\paragraph{Data} We split PSAT documents into 50\% train, 30\% test, and 20\% validation (the test and validation sets constitute 50\% of the data, such that the number of test and validation documents will be large enough). We then use the manual alignments in these document sets for training/testing/validation examples. Table~\ref{tab:data} shows the number of documents and aligned sentences in each split.

\paragraph{Evaluation}
For automatic evaluation, we report the following metrics: 

(1) \textbf{SARI}~\cite{xu2016optimizing}, a widely used metric especially designed according to the edit nature of simplification. We report the average n-gram F1 scores corresponding to keep, delete, and add operations.\footnote{Note that SARI scores are more reliable if multiple reference simplifications are present, but we are limited to one reference due to personnel, time, and budget constraints.}

(2) \textbf{BLEU}~\cite{papineni2002bleu}, a standard metric in machine translation and other conditional generation tasks. \citet{xu2016optimizing} stated that while SARI is better at capturing simplicity, BLEU has a stronger correlation with grammar and meaning.

(3) \textbf{BERTScore}~\cite{zhang2019bertscore}, which is shown to correlate better with human references in generation tasks, and has been adopted as an evaluation metric for summarization~\cite{ahuja2022aspectnews}.

\begin{table}
\centering
\small
\begin{tabular}{rccc}
\toprule
 & BERTScore & BLEU & SARI   \\ 
\midrule
ACCESS & 0.898 & 0.104 & \textbf{0.271}  \\ 
\midrule
T5-newsela & 0.906 & 0.132 & 0.226  \\ 
T5-wiki & 0.903 & 0.171 & 0.169  \\
\midrule
T5-newsela-ft & 0.919 & 0.210 & 0.241  \\ 
T5-wiki-ft & \textbf{0.923} & \textbf{0.239} & 0.216  \\ \bottomrule
\end{tabular}
    \caption{Automatic evaluation results of simplification systems.}
    \label{tab:results}
\end{table}

\subsection{Results}
Results are shown in Table~\ref{tab:results}. 
First, notice that all scores are lower in terms of absolute scale compared to the datasets they are trained on; for instance, \citet{devaraj2022evaluating} reported SARI scores over 36 for all models on Newsela and Wikilarge. 
Although \texttt{ACCESS} obtained the highest SARI score, its BLEU and BERTScores are the lowest; our manual inspection (Section~\ref{sec:humaneval}) confirmed that this model does not produce satisfactory simplification for admissions instructions. 
When comparing \texttt{T5-newsela} and \texttt{T5-wiki}, there is no clear superior model as \texttt{T5-wiki} has higher BLEU but lower SARI, and the two have very little difference in terms of BERTScore. 
This suggests that training on either Wikipedia articles or news articles as a starting point is acceptable.

The clearest trends across all metrics is that fine-tuning on PSAT substantially improves performance regardless of whether the model is pre-trained on Wikilarge or Newsela, demonstrating that in-domain fine-tuning is useful even with a moderately sized dataset like PSAT. This finding is consistent with recent discoveries in summarization~\cite{yu2021adaptsum}.

\begin{figure}[!t]
\noindent\includegraphics[width=\linewidth]{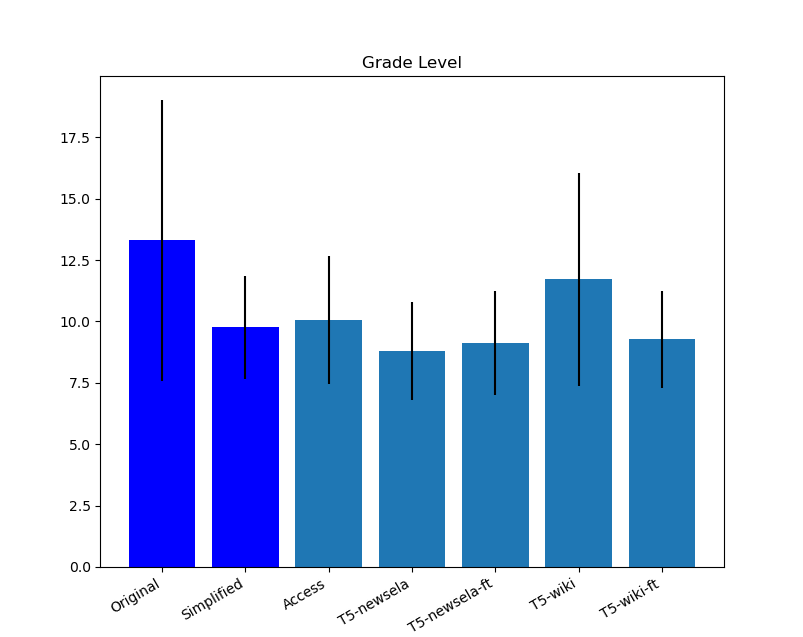}
\caption{The Flesch-Kincaid Grade Level averages and standard deviation for the original and manually simplified texts, as well as the simplified texts generated by the five models.}
\label{fig:modelfk}
\end{figure}

Figure~\ref{fig:modelfk} depicts the Flesch-Kincaid grade level for the original sentences, their manually simplified versions, and the model outputs among the aligned data. 
Note that although we perform this analysis due to Flesch-Kincaid's superior performance in assessing the readability of other technical texts in an unsupervised manner~\cite{devaraj2021paragraph}, we do not intend to use Flesch-Kincaid as one our main automatic evaluation metrics for reasons pointed out by~\citet{alva2021suitability} and \citet{tanprasert2021flesch}. 
Comparing the five models, \texttt{T5-wiki} yielded a higher (and more varied reading level) than all other models and the manually simplified. On the contrary, \texttt{T5-newsela}'s output reading level is slightly lower than the manual simplification.
Meanwhile, 
both \texttt{T5-newsela-ft} and \texttt{T5-wiki-ft} produced desirable readability levels; notably fine-tuning was able to improve the readability of \texttt{T5-wiki} outputs.

\section{Manual Inspection}\label{sec:humaneval}

Finally, because PSAT is a very new domain that comes with its own characteristics and requirements for accurate simplification, we perform a small scale manual inspection with the professional who authored the simplified documents; this professional has not seen model-generated outputs in NLP before. Because of time commitment required to inspect outputs from all 5 models, we were only able to study 25 examples. Nonetheless we hope the insights here will guide future work on domain-specific evaluation metrics.

Concretely, we randomly sample 25 sentences from the test set. We ask the professional to inspect outputs
based on three metrics: \emph{\bf fluency, simplicity, and accuracy}. Under each metric, we ask the professional to choose the highest ranked simplification; an option of \emph{none} is available if no model provided satisfactory outputs for any criteria. 

\begin{table}
\centering
\small
\begin{tabular}{rccc}
\toprule
 & Fluency & Simplicity & Accuracy \\ 
\midrule
ACCESS & 1/25 & 1/25 & 1/25  \\ \midrule
T5-newsela & 4/25 & 4/25 & 4/25 \\ 
T5-wiki & 4/25 & 4/25 & 5/25 \\
 \midrule
T5-newsela-ft & 2/25 & 2/25 & 1/25 \\
T5-wiki-ft & 5/25 & 5/25 & 5/25  \\ \midrule
None & 9/25 & 9/25 & 9/25 \\ \bottomrule
\end{tabular}
    \caption{Number of times (out of 25) that each system received the highest rank, along with \# of times that no system was selected.}
    \label{tab:humaneval}
\end{table}

Results in Table~\ref{tab:humaneval} confirmed that \texttt{T5-wiki-ft} performed the best, and that \texttt{ACCESS} outputs do not outperform other models. Similar to automatic evaluation results, \texttt{T5-wiki} vs \texttt{T5-newsela}  However, there is also a large portion of examples where ``none'' was chosen. Additionally, \texttt{T5-newsela-ft} performed much worse than \texttt{T5-newsela}. This suggests that automatic metrics do not capture some of the aspects that the professional deems important in their judgments. This echos findings from~\citet{alva2021suitability} but emphasizes the importance of human evaluation for specific domains. We detail some of the insights qualitatively below.

First, several simplifications required adding words to the simplified version that did not appear in the original, however the model did not learn to do so.
The most frequent example was simplifications adding ``official'' to the phrase ``Official Transcript(s)''. Here, subject-matter experts felt that adding ``official'' was necessary to make it clearer and more simple for students to understand that they needed to submit official transcripts instead of copies or screenshots of their transcripts. 

Similarly, many simplifications added the subject and modal verb ``you must'' to phrases such as ``submit your official transcript(s)'', ``complete an application'', and ``apply by the priority date'', as discussed in Section~\ref{sec:simplification_process}. 
The subject-matter experts felt it was necessary to clarify that a student ``must'' complete these tasks, otherwise they would not be considered for admission. 
However, the model could not generate such phrasing.

Finally, many models removed keywords from the original versions.
This most often happened when an institution accepted multiple application types, such as the Coalition for College Application and the Common Application. For instance, one sentence original sentence read, ``\emph{We accept the Coalition for College Application and the Common Application}''. However, several models attempted to simplify the sentence and remove either the Coalition for College Application or Common Application, as they both contain the word ``application'', and the model may have removed these ``redundant'' phrases. However, both application types needed to appear in the simplified version to gain subject-matter expert approval, as an accurate admissions instructions contain all of the different applications that a student may use to apply for admission, not just the simplest application. As a result, several models did not accurately keep keywords from original to simplified versions because the models did not have this domain-specific information that subject-matter experts had.

\section{Discussions and Conclusions}

This work presents PSAT, a text simplification corpus consisting of admission instructions texts from 112 US higher education institutions and their simplified versions. PSAT texts are professionally simplified and verified, rendering it the first-of-its-kind and most accurate dataset in this high-stake domain. We showed that this dataset is challenging for existing simplification models, especially due to domain mismatch, and domain-specific requirements for the accuracy of information.

We recognize the limitations of this study. While there are over 6,000 institutions of higher education in the US, this dataset sampled a small number of these institutions given time constraints and the substantial work necessary to gather and simplify text and work with subject-matter experts to approve simplified texts. 

Ultimately, we will perform future research—using this dataset—to specifically identify which simplifications were acceptable and unacceptable (at the lexical item-, sentence- and paragraph-level), thus informing admissions practitioners as to how text can or cannot be simplified for students and prospective students, broadly. We hope PSAT is a first step towards automatic simplification systems for fairer access to higher education.  

\section*{Acknowledgements}
The authors would like to acknowledge the subject-matter experts who volunteered their time to read the texts from this study and provide their professional judgement. 
This research is partially supported by NSF grants IIS-2145479, IIS-2107524. We acknowledge the Texas Advanced Computing Center (TACC)\footnote{\url{https://www.tacc.utexas.edu}} at UT Austin for many of the results within this paper. 

% Entries for the entire Anthology, followed by custom entries
\bibliography{anthology}

\begin{thebibliography}{67}
\expandafter\ifx\csname natexlab\endcsname\relax\def\natexlab#1{#1}\fi

\bibitem[{Ahuja et~al.(2022)Ahuja, Xu, Gupta, Horecka, and
  Durrett}]{ahuja2022aspectnews}
Ojas Ahuja, Jiacheng Xu, Akshay Gupta, Kevin Horecka, and Greg Durrett. 2022.
\newblock {ASPECTNEWS}: Aspect-oriented summarization of news documents.
\newblock In \emph{Proceedings of the 60th Annual Meeting of the Association
  for Computational Linguistics (Volume 1: Long Papers)}, pages 6494--6506.

\bibitem[{Alva-Manchego et~al.(2020)Alva-Manchego, Scarton, and
  Specia}]{alva2020data}
Fernando Alva-Manchego, Carolina Scarton, and Lucia Specia. 2020.
\newblock Data-driven sentence simplification: Survey and benchmark.
\newblock \emph{Computational Linguistics}, 46(1):135--187.

\bibitem[{Alva-Manchego et~al.(2021)Alva-Manchego, Scarton, and
  Specia}]{alva2021suitability}
Fernando Alva-Manchego, Carolina Scarton, and Lucia Specia. 2021.
\newblock The (un) suitability of automatic evaluation metrics for text
  simplification.
\newblock \emph{Computational Linguistics}, 47(4):861--889.

\bibitem[{Ardoin(2013)}]{ardoin2013learning}
Mary~Sonja Ardoin. 2013.
\newblock \emph{Learning a different language: Rural students' comprehension of
  college knowledge and university jargon}.
\newblock North Carolina State University.

\bibitem[{Astin et~al.(1982)}]{astin1982minorities}
Alexander~W Astin et~al. 1982.
\newblock \emph{Minorities in American higher education. Recent trends, current
  prospects, and recommendations}.
\newblock ERIC.

\bibitem[{Auerbach(2004)}]{auerbach2004engaging}
Susan Auerbach. 2004.
\newblock Engaging latino parents in supporting college pathways: Lessons from
  a college access program.
\newblock \emph{Journal of Hispanic Higher Education}, 3(2):125--145.

\bibitem[{Cannon(1989)}]{cannon1989abbreviations}
Garland Cannon. 1989.
\newblock Abbreviations and acronyms in english word-formation.
\newblock \emph{American Speech}, 64(2):99--127.

\bibitem[{Coleman(1962)}]{coleman1962improving}
Edmund~Benedict Coleman. 1962.
\newblock Improving comprehensibility by shortening sentences.
\newblock \emph{Journal of Applied Psychology}, 46(2):131.

\bibitem[{Cook et~al.(2012)Cook, P{\'e}russe, and Rojas}]{cook2012increasing}
Amy Cook, Rachelle P{\'e}russe, and Eliana~D Rojas. 2012.
\newblock Increasing academic achievement and college-going rates for latina/o
  english language learners: A survey of school counselor interventions.
\newblock \emph{The Journal of Counselor Preparation and Supervision}, 4(2):2.

\bibitem[{Coster and Kauchak(2011)}]{coster2011simple}
William Coster and David Kauchak. 2011.
\newblock Simple {E}nglish {W}ikipedia: a new text simplification task.
\newblock In \emph{Proceedings of the 49th Annual Meeting of the Association
  for Computational Linguistics: Human Language Technologies}, pages 665--669.

\bibitem[{Crossley et~al.(2008)Crossley, Greenfield, and
  Mcnamara}]{crossley2008text}
SA~Crossley, J~Greenfield, and DS~Mcnamara. 2008.
\newblock Text readability using assessing based indices.
\newblock \emph{TESOL Quarterly}, 42(3):475--493.

\bibitem[{Devaraj et~al.(2021)Devaraj, Marshall, Wallace, and
  Li}]{devaraj2021paragraph}
Ashwin Devaraj, Iain Marshall, Byron~C Wallace, and Junyi~Jessy Li. 2021.
\newblock Paragraph-level simplification of medical texts.
\newblock In \emph{Proceedings of the 2021 Conference of the North American
  Chapter of the Association for Computational Linguistics: Human Language
  Technologies}, pages 4972--4984.

\bibitem[{Devaraj et~al.(2022)Devaraj, Sheffield, Wallace, and
  Li}]{devaraj2022evaluating}
Ashwin Devaraj, William Sheffield, Byron Wallace, and Junyi~Jessy Li. 2022.
\newblock Evaluating factuality in text simplification.
\newblock In \emph{Proceedings of the 60th Annual Meeting of the Association
  for Computational Linguistics (Volume 1: Long Papers)}, pages 7331--7345.

\bibitem[{DeVito(1969)}]{devito1969some}
Joseph~A DeVito. 1969.
\newblock Some psycholinguistic aspects of active and passive sentences.
\newblock \emph{Quarterly Journal of Speech}, 55(4):401--406.

\bibitem[{Dong et~al.(2019)Dong, Li, Rezagholizadeh, and
  Cheung}]{dong2019editnts}
Yue Dong, Zichao Li, Mehdi Rezagholizadeh, and Jackie Chi~Kit Cheung. 2019.
\newblock {EditNTS}: An neural programmer-interpreter model for sentence
  simplification through explicit editing.
\newblock In \emph{Proceedings of the 57th Annual Meeting of the Association
  for Computational Linguistics}, pages 3393--3402.

\bibitem[{Ferreira(1994)}]{ferreira1994choice}
Fernanda Ferreira. 1994.
\newblock Choice of passive voice is affected by verb type and animacy.
\newblock \emph{Journal of Memory and Language}, 33(6):715--736.

\bibitem[{Flores(2010)}]{flores2010state}
Stella~M Flores. 2010.
\newblock State dream acts: The effect of in-state resident tuition policies
  and undocumented latino students.
\newblock \emph{The Review of Higher Education}, 33(2):239--283.

\bibitem[{Gandara(1986)}]{gandara1986chicanos}
Patricia Gandara. 1986.
\newblock Chicanos in higher education: The politics of self-interest.
\newblock \emph{American Journal of Education}, 95(1):256--272.

\bibitem[{Goff et~al.(2004)Goff, Patino, and Jackson}]{goff2004preferred}
Brent Goff, Vanessa Patino, and Gary Jackson. 2004.
\newblock Preferred information sources of high school students for community
  colleges and universities.
\newblock \emph{Community College Journal of Research \& Practice},
  28(10):795--803.

\bibitem[{Hartman(1997)}]{hartman1997college}
Kenneth~E Hartman. 1997.
\newblock College selection and the internet: Advice for schools, a wake-up
  call for colleges.
\newblock \emph{Journal of College Admission}, 154:22--31.

\bibitem[{Hauser et~al.(2005)Hauser, Edley~Jr, Koenig, and
  Elliott}]{hauser2005measuring}
Robert~M Hauser, Christopher~F Edley~Jr, Judith~Anderson Koenig, and Stuart~W
  Elliott. 2005.
\newblock \emph{Measuring Literacy: Performance Levels for Adults.}
\newblock ERIC.

\bibitem[{Hossler and Gallagher(1987)}]{hossler1987studying}
Don Hossler and Karen Gallagher. 1987.
\newblock Studying student college choice: A three-phase model and the
  implication...-supersearch powered by summon.
\newblock \emph{College and University}, 62:201--21.

\bibitem[{Hulme et~al.(1997)Hulme, Roodenrys, Schweickert, Brown, Martin, and
  Stuart}]{hulme1997word}
Charles Hulme, Steven Roodenrys, Richard Schweickert, Gordon~DA Brown, Sarah
  Martin, and George Stuart. 1997.
\newblock Word-frequency effects on short-term memory tasks: evidence for a
  redintegration process in immediate serial recall.
\newblock \emph{Journal of Experimental Psychology: Learning, Memory, and
  Cognition}, 23(5):1217.

\bibitem[{Hurtado et~al.(2020)Hurtado, Ramos, Perez, and
  Lopez-Salgado}]{hurtado2020latinx}
Sylvia Hurtado, Hector~Vicente Ramos, Edwin Perez, and Xochilth Lopez-Salgado.
  2020.
\newblock Latinx student assets, college readiness, and access: Are we making
  progress?
\newblock \emph{Education Sciences}, 10(4):100.

\bibitem[{Ibrahim(1989)}]{ibrahim1989acronyms}
AM~Ibrahim. 1989.
\newblock Acronyms observed.
\newblock \emph{IEEE Transactions on Professional Communication}, 32(1):27--28.

\bibitem[{Javourey-Drevet et~al.(2022)Javourey-Drevet, Dufau, Fran{\c{c}}ois,
  Gala, Ginesti{\'e}, and Ziegler}]{javourey2022simplification}
Ludivine Javourey-Drevet, St{\'e}phane Dufau, Thomas Fran{\c{c}}ois, N{\'u}ria
  Gala, Jacques Ginesti{\'e}, and Johannes~C Ziegler. 2022.
\newblock Simplification of literary and scientific texts to improve reading
  fluency and comprehension in beginning readers of french.
\newblock \emph{Applied Psycholinguistics}, 43(2):485--512.

\bibitem[{Jiang et~al.(2020)Jiang, Maddela, Lan, Zhong, and
  Xu}]{jiang2020neural}
Chao Jiang, Mounica Maddela, Wuwei Lan, Yang Zhong, and Wei Xu. 2020.
\newblock Neural {CRF} model for sentence alignment in text simplification.
\newblock In \emph{Proceedings of the 58th Annual Meeting of the Association
  for Computational Linguistics}, pages 7943--7960.

\bibitem[{Kanno(2018)}]{kanno2018non}
Yasuko Kanno. 2018.
\newblock Non-college-bound english learners as the underserved third: How
  students graduate from high school neither college-nor career-ready.
\newblock \emph{Journal of Education for Students Placed at Risk (JESPAR)},
  23(4):336--358.

\bibitem[{Kincaid et~al.(1975)Kincaid, Fishburne~Jr, Rogers, and
  Chissom}]{kincaid1975derivation}
J~Peter Kincaid, Robert~P Fishburne~Jr, Richard~L Rogers, and Brad~S Chissom.
  1975.
\newblock Derivation of new readability formulas (automated readability index,
  fog count and flesch reading ease formula) for navy enlisted personnel.
\newblock Technical report, Naval Technical Training Command Millington TN
  Research Branch.

\bibitem[{Kriz et~al.(2019)Kriz, Sedoc, Apidianaki, Zheng, Kumar, Miltsakaki,
  and Callison-Burch}]{kriz2019complexity}
Reno Kriz, Jo{\~a}o Sedoc, Marianna Apidianaki, Carolina Zheng, Gaurav Kumar,
  Eleni Miltsakaki, and Chris Callison-Burch. 2019.
\newblock Complexity-weighted loss and diverse reranking for sentence
  simplification.
\newblock In \emph{Proceedings of the 2019 Conference of the North American
  Chapter of the Association for Computational Linguistics: Human Language
  Technologies, Volume 1 (Long and Short Papers)}, pages 3137--3147.

\bibitem[{Laszlo and Federmeier(2007)}]{laszlo2007better}
Sarah Laszlo and Kara~D Federmeier. 2007.
\newblock Better the dvl you know: Acronyms reveal the contribution of
  familiarity to single-word reading.
\newblock \emph{Psychological Science}, 18(2):122--126.

\bibitem[{Maddela et~al.(2021)Maddela, Alva-Manchego, and
  Xu}]{maddela2021controllable}
Mounica Maddela, Fernando Alva-Manchego, and Wei Xu. 2021.
\newblock Controllable text simplification with explicit paraphrasing.
\newblock In \emph{Proceedings of the 2021 Conference of the North American
  Chapter of the Association for Computational Linguistics: Human Language
  Technologies}, pages 3536--3553.

\bibitem[{Mamedova and Pawlowski(2019)}]{mamedova2019adult}
Saida Mamedova and Emily Pawlowski. 2019.
\newblock Adult literacy in the united states. data point. nces 2019-179.
\newblock \emph{National Center for Education Statistics}.

\bibitem[{Martin et~al.(2020)Martin, De~La~Clergerie, Sagot, and
  Bordes}]{martin2020controllable}
Louis Martin, {\'E}ric~Villemonte De~La~Clergerie, Beno{\^\i}t Sagot, and
  Antoine Bordes. 2020.
\newblock Controllable sentence simplification.
\newblock In \emph{Proceedings of the 12th Language Resources and Evaluation
  Conference}, pages 4689--4698.

\bibitem[{McBrien(2005)}]{mcbrien2005educational}
J~Lynn McBrien. 2005.
\newblock Educational needs and barriers for refugee students in the united
  states: A review of the literature.
\newblock \emph{Review of educational research}, 75(3):329--364.

\bibitem[{McCulloh(2022)}]{mcculloh2022exploration}
Edna McCulloh. 2022.
\newblock An exploration of parental support in the retention of rural
  first-generation college students.
\newblock \emph{Journal of College Student Retention: Research, Theory \&
  Practice}, 24(1):144--168.

\bibitem[{McNamara et~al.(2014)McNamara, Graesser, McCarthy, and
  Cai}]{mcnamara2014automated}
Danielle~S McNamara, Arthur~C Graesser, Philip~M McCarthy, and Zhiqiang Cai.
  2014.
\newblock \emph{Automated evaluation of text and discourse with Coh-Metrix}.
\newblock Cambridge University Press.

\bibitem[{Monaghan et~al.(2017)Monaghan, Chang, Welbourne, and
  Brysbaert}]{monaghan2017exploring}
Padraic Monaghan, Ya-Ning Chang, Stephen Welbourne, and Marc Brysbaert. 2017.
\newblock Exploring the relations between word frequency, language exposure,
  and bilingualism in a computational model of reading.
\newblock \emph{Journal of Memory and Language}, 93:1--21.

\bibitem[{Morrison(2021)}]{morrison_2021}
Joe Morrison. 2021.
\newblock \href
  {https://www.eschoolnews.com/2021/03/22/4-ways-to-transform-college-admissions/}
  {4 ways to transform college admissions}.
\newblock \emph{eSchool News}.

\bibitem[{Nye and Nenkova(2015)}]{nye2015identification}
Benjamin Nye and Ani Nenkova. 2015.
\newblock Identification and characterization of newsworthy verbs in world
  news.
\newblock In \emph{Proceedings of the 2015 Conference of the North American
  Chapter of the Association for Computational Linguistics: Human Language
  Technologies}, pages 1440--1445.

\bibitem[{Papineni et~al.(2002)Papineni, Roukos, Ward, and
  Zhu}]{papineni2002bleu}
Kishore Papineni, Salim Roukos, Todd Ward, and Wei-Jing Zhu. 2002.
\newblock Bleu: a method for automatic evaluation of machine translation.
\newblock In \emph{Proceedings of the 40th annual meeting of the Association
  for Computational Linguistics}, pages 311--318.

\bibitem[{P{\'e}rez and McDonough(2008)}]{perez2008journal}
Patricia~A P{\'e}rez and Patricia~M McDonough. 2008.
\newblock Journal of hispanic higher.
\newblock \emph{Journal of Hispanic Higher Education}, 7:249.

\bibitem[{Post(1990)}]{post1990college}
David Post. 1990.
\newblock College-going decisions by chicanos: The politics of misinformation.
\newblock \emph{Educational Evaluation and Policy Analysis}, 12(2):174--187.

\bibitem[{Raffel et~al.(2020)Raffel, Shazeer, Roberts, Lee, Narang, Matena,
  Zhou, Li, and Liu}]{raffel2020exploring}
Colin Raffel, Noam Shazeer, Adam Roberts, Katherine Lee, Sharan Narang, Michael
  Matena, Yanqi Zhou, Wei Li, and Peter~J Liu. 2020.
\newblock Exploring the limits of transfer learning with a unified text-to-text
  transformer.
\newblock \emph{Journal of Machine Learning Research}, 21:1--67.

\bibitem[{Reilly(2021)}]{reilly_2021}
Katie Reilly. 2021.
\newblock \href {https://time.com/5951236/college-applications-covid-19/}
  {Applying to college was hard. covid-19 made it impossible}.
\newblock \emph{Time}.

\bibitem[{Rosa(2006)}]{rosa2006opportunity}
Mari Luna De~La Rosa. 2006.
\newblock Is opportunity knocking? low-income students’ perceptions of
  college and financial aid.
\newblock \emph{American behavioral scientist}, 49(12):1670--1686.

\bibitem[{R{\'u}a(2002)}]{rua2002structure}
Paula~L{\'o}pez R{\'u}a. 2002.
\newblock On the structure of acronyms and neighbouring categories: A
  prototype-based account.
\newblock \emph{English Language \& Linguistics}, 6(1):31--60.

\bibitem[{Siddharthan(2014)}]{siddharthan2014survey}
Advaith Siddharthan. 2014.
\newblock A survey of research on text simplification.
\newblock \emph{ITL-International Journal of Applied Linguistics},
  165(2):259--298.

\bibitem[{Sum et~al.(2004)Sum, Kirsch, and Yamamoto}]{sum2004pathways}
Andrew Sum, Irwin Kirsch, and Kentaro Yamamoto. 2004.
\newblock Pathways to labor market success: The literacy proficiency of {US}
  adults.
\newblock \emph{Policy Information Report, Educational Testing Service}.

\bibitem[{Taghva and Gilbreth(1999)}]{taghva1999recognizing}
Kazem Taghva and Jeff Gilbreth. 1999.
\newblock Recognizing acronyms and their definitions.
\newblock \emph{International Journal on Document Analysis and Recognition},
  1(4):191--198.

\bibitem[{Tanprasert and Kauchak(2021)}]{tanprasert2021flesch}
Teerapaun Tanprasert and David Kauchak. 2021.
\newblock {F}lesch-{K}incaid is not a text simplification evaluation metric.
\newblock In \emph{Proceedings of the 1st Workshop on Natural Language
  Generation, Evaluation, and Metrics (GEM 2021)}, pages 1--14.

\bibitem[{Taylor(2017)}]{taylor2017twenty}
Zachary~W Taylor. 2017.
\newblock Twenty-first-century communicators: How registrars can reject
  ‘university idiolect’.
\newblock \emph{The Successful Registrar}, 17(9):6--6.

\bibitem[{Taylor(2018)}]{taylor2018intelligibility}
Zachary~W Taylor. 2018.
\newblock Intelligibility is equity: Can international students read
  undergraduate admissions materials?
\newblock \emph{Higher Education Quarterly}, 72(2):160--169.

\bibitem[{Taylor(2019)}]{taylor2019lost}
Zachary~W Taylor. 2019.
\newblock Lost in translation: Linguistic hurdles facing us language-minority
  students of color pursuing international higher education.
\newblock \emph{. Lost in Translation: Linguistic Hurdles Facing US
  Language-Minority Students of Color Pursuing International Higher Education.
  Emerging Voices in Education}, 1(1):45--60.

\bibitem[{Taylor(2020{\natexlab{a}})}]{taylor2020comprenderan}
Zachary~W Taylor. 2020{\natexlab{a}}.
\newblock ?` comprender{\'a}n mis amigos y la familia? analyzing spanish
  translations of admission materials for latina/o students applying to 4-year
  institutions in the united states.
\newblock \emph{Journal of Hispanic Higher Education}, 19(2):195--209.

\bibitem[{Taylor(2020{\natexlab{b}})}]{taylor2020college}
Zachary~W Taylor. 2020{\natexlab{b}}.
\newblock College admissions for l2 students: Comparing l1 and l2 readability
  of admissions materials for us higher education.
\newblock \emph{Journal of College Access (2020)}.

\bibitem[{Taylor and Bicak(2019)}]{taylor2019fafsa}
Zachary~W Taylor and Ibrahim Bicak. 2019.
\newblock What is the {FAFSA}? an adult learner knowledge survey of student
  financial aid jargon.
\newblock \emph{Journal of Adult and Continuing Education}, 25(1):94--112.

\bibitem[{Tornatzky et~al.(2002)Tornatzky, Cutler, and
  Lee}]{tornatzky2002college}
Louis~G Tornatzky, Richard Cutler, and Jongho Lee. 2002.
\newblock College knowledge: What latino parents need to know and why they
  don't know it.
\newblock \emph{Joyce Foundation}.

\bibitem[{Wang et~al.(2016)Wang, Chen, Rochford, and Qiang}]{wang2016text}
Tong Wang, Ping Chen, John Rochford, and Jipeng Qiang. 2016.
\newblock Text simplification using neural machine translation.
\newblock In \emph{Thirtieth AAAI Conference on Artificial Intelligence}.

\bibitem[{Ward(2006)}]{ward2006improving}
Nadia~L Ward. 2006.
\newblock Improving equity and access for low-income and minority youth into
  institutions of higher education.
\newblock \emph{Urban Education}, 41(1):50--70.

\bibitem[{Woodsend and Lapata(2011)}]{woodsend2011wikisimple}
Kristian Woodsend and Mirella Lapata. 2011.
\newblock Wikisimple: Automatic simplification of wikipedia articles.
\newblock In \emph{Twenty-fifth AAAI conference on artificial intelligence}.

\bibitem[{Xu et~al.(2015)Xu, Callison-Burch, and Napoles}]{xu2015problems}
Wei Xu, Chris Callison-Burch, and Courtney Napoles. 2015.
\newblock Problems in current text simplification research: New data can help.
\newblock \emph{Transactions of the Association for Computational Linguistics},
  3:283--297.

\bibitem[{Xu et~al.(2016)Xu, Napoles, Pavlick, Chen, and
  Callison-Burch}]{xu2016optimizing}
Wei Xu, Courtney Napoles, Ellie Pavlick, Quanze Chen, and Chris Callison-Burch.
  2016.
\newblock Optimizing statistical machine translation for text simplification.
\newblock \emph{Transactions of the Association for Computational Linguistics},
  4:401--415.

\bibitem[{Yano et~al.(1994)Yano, Long, and Ross}]{yano1994effects}
Yasukata Yano, Michael~H Long, and Steven Ross. 1994.
\newblock The effects of simplified and elaborated texts on foreign language
  reading comprehension.
\newblock \emph{Language learning}, 44(2):189--219.

\bibitem[{Yu et~al.(2021)Yu, Liu, and Fung}]{yu2021adaptsum}
Tiezheng Yu, Zihan Liu, and Pascale Fung. 2021.
\newblock Adaptsum: Towards low-resource domain adaptation for abstractive
  summarization.
\newblock In \emph{Proceedings of the 2021 Conference of the North American
  Chapter of the Association for Computational Linguistics: Human Language
  Technologies}, pages 5892--5904.

\bibitem[{Zhang et~al.(2019)Zhang, Kishore, Wu, Weinberger, and
  Artzi}]{zhang2019bertscore}
Tianyi Zhang, Varsha Kishore, Felix Wu, Kilian~Q Weinberger, and Yoav Artzi.
  2019.
\newblock Bertscore: Evaluating text generation with bert.
\newblock In \emph{International Conference on Learning Representations}.

\bibitem[{Zhang and Lapata(2017)}]{zhang2017sentence}
Xingxing Zhang and Mirella Lapata. 2017.
\newblock Sentence simplification with deep reinforcement learning.
\newblock In \emph{Proceedings of the 2017 Conference on Empirical Methods in
  Natural Language Processing}, pages 584--594.

\end{thebibliography}

%\clearpage

%\appendix

%\section{Example Simplification}
%\label{sec:appendix}

%This is an appendix.

\end{document}